\documentclass[conference]{IEEEtran}
\IEEEoverridecommandlockouts
\usepackage{cite}
\usepackage{amsmath,amssymb,amsfonts}
\usepackage{algorithmic}
\usepackage{graphicx}
\usepackage{array}
\usepackage{float}
\usepackage{subcaption}
\usepackage{fancyhdr}

\usepackage{microtype}  % Improves overall text appearance
\usepackage{multirow}
\usepackage{textcomp}
\usepackage{xcolor}
\usepackage{censor}

\def\BibTeX{{\rm B\kern-.05em{\sc i\kern-.025em b}\kern-.08em
    T\kern-.1667em\lower.7ex\hbox{E}\kern-.125emX}}

\begin{document}

\title{Explainable Multimodal Sentiment Analysis on Bengali Memes\\}

\author{
    \IEEEauthorblockN{Kazi Toufique Elahi, Tasnuva Binte Rahman, Shakil Shahriar, Samir Sarker,\\ Sajib Kumar Saha Joy, Faisal Muhammad Shah}
    \IEEEauthorblockA{Department of Computer Science and Engineering\\
    Ahsanullah University of Science and Technology \\
    Dhaka 1208, Bangladesh\\
    Email: \{ktoufiquee, tasnuvabinterahmansrishti, shakilshahriararnob, rohitsarker5\}@gmail.com,\\\{joy.cse, faisal.cse\}@aust.edu}
}

\maketitle

\begin{abstract}
Memes have become a distinctive and effective form of communication in the digital era, attracting online communities and cutting across cultural barriers. Even though memes are frequently linked with humor, they have an amazing capacity to convey a wide range of emotions, including happiness, sarcasm, frustration, and more. Understanding and interpreting the sentiment underlying memes has become crucial in the age of information. Previous research has explored text-based, image-based, and multimodal approaches, leading to the development of models like CAPSAN and PromptHate for detecting various meme categories. However, the study of low-resource languages like Bengali memes remains scarce, with limited availability of publicly accessible datasets. A recent contribution includes the introduction of the MemoSen dataset. However, the achieved accuracy is notably low, and the dataset suffers from imbalanced distribution. In this study, we employed a multimodal approach using ResNet50 and BanglishBERT and achieved a satisfactory result of 0.71 weighted F1-score, performed comparison with unimodal approaches, and interpreted behaviors of the models using explainable artificial intelligence (XAI) techniques.
\end{abstract}


\section{Introduction}
The usage of social media platforms (i.e. Facebook, Twitter, Instagram) has increased dramatically due to the substantial evolution of the Internet and various Web 2.0 applications \cite{hossain2022memosen}. These social media platforms contain people's opinions and conversations on various topics like politics, games, entertainment, and so on. Sentiment analysis on these has always been a popular topic for NLP researchers. However, the nature of data found on these platforms has evolved over time. It has transformed from being solely text-based to encompass a combination of text, images, audio, and videos. One such content is memes. Memes are a popular form of communication amongst the younger generation. It is a type of content that is context-based and might be in the form of images, text, or both. Memes are entertaining, critical, sarcastic, and may even be political \cite{asmawati2022sentiment}. Movies, politics, TV shows, religion, and social culture are all common sources of Internet memes \cite{alluri2021multi}. As it can be expressed in a lot of ways and can express a wide range of emotions, understanding and interpreting the sentiment behind memes is of utmost relevance in this age.\par
Several research studies have been conducted to better understand the sentiment underlying memes. Various datasets specifically curated for memes, such as Facebook Hateful Memes (FHM) \cite{kiela2020hateful}, Harmful Meme Dataset (HarM) \cite{pramanick2021detecting}, and Memotion \cite{chhavi2020memotion}, are available publicly. These datasets have served as a foundation for several significant researches. For instance, Alluri et al. \cite{alluri2021multi} introduced the CAPSEN model with an accuracy of 60.25\% (Memotion). Similarly, Cao et al. \cite{cao2023prompting} introduced a prompt-based model PromptHate with an accuracy of 84.47\% (FHM) and 72.98\% (HarM).\par 
However, it is essential to highlight that most of these studies are focused on datasets in the English language. As Bengali is a low-resource language, the availability of datasets for memes in the Bengali language is scarce, and research in this domain is even rarer. Nonetheless, a recent study conducted by Hossain et al. \cite{hossain2022memosen} addressed this gap by introducing the MemoSen dataset, comprising a collection of Bengali memes for sentiment analysis. While their comprehensive analysis of the dataset yielded a weighted F1 score of 0.643, it remains a far cry from the scores achieved in sentiment analysis of memes in other languages. Additionally, the dataset suffers from an inherent imbalance in its distribution. \par
Due to the limited availability of research on sentiment
analysis of Bengali memes, we aim to address
this gap by conducting research on the MemoSEN dataset.
Initially, we implemented different models using solely text inputs and image inputs that attained the highest F1-score of 0.66 and 0.70 respectively. Afterward, we combined both text and image inputs to create a multimodal approach that achieved an F1-score of 0.71. In summary, we have achieved the following: 
\begin{itemize} 
\item Reached a desirable accuracy by implementing a multimodal approach on the MemoSEN dataset. 
\item Evaluated the performance of our approach by performing a comparative analysis against the performance of other models. 
\item Leveraged Explainable AI (XAI) to provide an explanation of the implemented models.
\end{itemize}

\section{Related Works}
Numerous studies have been conducted on sentiment analysis of memes, employing both unimodal and multimodal approaches. A short description of studies that achieved exemplary performance is listed in this section.\par
Hossain et al.\cite{hossain2022memosen} introduced MemoSen, one of the largest publicly available multiclass sentiment classification datasets for Bengali memes. Their ResNet50+BiLSTM-based model achieved the highest 0.635 weighted F1-score showing that the multimodal approach performed better than the unimodal approach. The same authors created another multimodal hateful meme dataset named MUTE \cite{hossain2022mute} more recently, which contains 4158 memes with Bengali and code-mix captions, categorized into Hate and Non-hate classes. For benchmarking purposes, VGG16, VGG19, and ResNet50 models were employed for visual data. Whereas CNN+BiLSTM, BiLSTM+Attention, M-BERT, Bangla-BERT, and Cross-Lingual Representation Learner (XLM-R) were implemented for textual data. The highest weighted F1-Score of 0.672 was achieved by a multimodal approach using VGG16+BanglaBERT. Noticeably, the model faced trouble in correctly labeling the same words in different textual formats (code-mixed, code-switched). Additionally, the model is confused by the disparity between some memes' visual and textual content.\par Karim et al. \cite{karim2022multimodal}  achieved a maximum F1-score of 0.83 for hate speech detection on Bengali memes implementing XLM-RoBERTa+DenseNet-161. Their limitations included limited training data and the need for explanations regarding the model's predictions. Jannat et al.\cite{jannat2022empirical} achieved the highest F1-score of 0.68 using BiLSTM+VGG19 for binary sentiment classification in Bengali memes. CNN, BiLSTM, and BiLSTM-CNN were implemented to perform text classification. VGG16, VGG19, and InceptionV3 had been employed for visual data. BiLSTM and VGG19 came out to be the best performers individually. Sun et al. \cite{sun2021two} proposed TIMF, a model based on CNN+BiLSTM and TensorFusion, and performed benchmark analysis on multiple datasets. \par
Alluri et al.\cite{varma2021multimodal} addressed the class imbalance in the Memotion dataset using the Synthetic Minority Over-sampling Technique (SMOTE) and introduced IMGTXT (ViT+RoBERTa), IMGSEN(ViT+SBERT RoBERTa), and CAPSEN(Generated captions + Textual data using SBERT) models. The IMGSEN model outperformed other models in humor and motivation categories with F1-scores of 63.30 and 55.09, while the CAPSEN model excelled in sarcasm and overall sentiment with F1-scores of 61.33 and 57.79. It was suggested that the high dimensionality of data contributed to lower accuracy in IMGTXT. Additionally, it was mentioned that the CAPSEN model can be improved as it was only trained on Flickr8k \cite{hodosh2013framing}. \par 
Hasan et al.\cite{hasan2022cuet} achieved the highest F1-score of 49\% using a CNN+VGG16-based multimodal model in their study on the detection of troll memes. Their model struggled to identify non-troll memes in particular, most likely due to the overlapping nature of memes across all classes. Pramanik et al.\cite{pramanick2021detecting} contributed to the prospect of harmful meme detection with their approach of using VisualBERT pre-trained on COCO \cite{lin2014microsoft}, achieving the highest F1-score of 80.13\%, leading by almost 9\% to their unimodal pre-trained approach of ConcatBERT. They also conducted a human evaluation to justify their result and the VisualBERT performed best there as well with an almost 66\% F1 score. Ullah et al.\cite{ullah2017overview} in their overview of multimodal sentiment analysis mentioned the detection of hidden emotions, exploring more feature points of the face, and fusing analysis results from multiple modalities as the difficulties mostly needed to be handled regarding the subject. \par

\section{Dataset Overview}

In this study, the MemoSEN\cite{hossain2022memosen} dataset is utilized as the primary data source. Presently, the MemoSEN dataset comprises a total of 4372 images, each accompanied by its respective caption. It is worth mentioning that among these images, 17 do not have any associated captions. The memes within this dataset are categorized into three distinct groups: \textit{Neutral}, \textit{Positive}, and \textit{Negative}.

\begin{figure}[h]
    \centering
    \begin{subfigure}[b]{0.15\textwidth}
        \includegraphics[width=\textwidth]{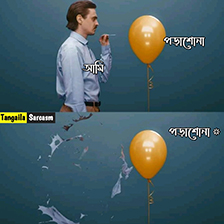}
        \label{fig:neu}
        \subcaption[]{Neutral}
    \end{subfigure}
    \begin{subfigure}[b]{0.15\textwidth}
        \includegraphics[width=\textwidth]{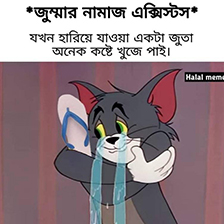}
        \label{fig:pos}
        \subcaption[]{Positive}
    \end{subfigure}
    \begin{subfigure}[b]{0.15\textwidth}
        \includegraphics[width=\textwidth]{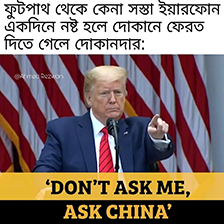}
        \label{fig:neg}
        \subcaption[]{Negative}
    \end{subfigure}
    \caption{Example meme from each class}
    \label{fig:eg}
\end{figure}

An example from each class is visualized in Figure \ref{fig:eg}. Through observation, it can be seen that the people who made this dataset labeled neutral to self-criticizing memes in general, most of the neutral memes are generally self-loathing, self-humor, or self-criticization.

\begin{table}[h]
\centering
\caption{Dataset Statistics}\label{dset}
\begin{tabular}{l|l|l|l|}
\cline{2-4}
                                        & Neutral & Positive & Negative \\ \hline
\multicolumn{1}{|l|}{Train (70\%)}      & 204     & 944      & 1909     \\ \hline
\multicolumn{1}{|l|}{Validation (10\%)} & 29      & 135      & 273      \\ \hline
\multicolumn{1}{|l|}{Test (20\%)}       & 58      & 270      & 546      \\ \hline
\multicolumn{1}{|l|}{Total}             & 291     & 1349     & 2728     \\ \hline
\end{tabular}
\end{table}

For our study, we stratified and split the dataset into train, validation, and test sets. The distribution of memes across the three categories is shown in Table \ref{dset}. It is evident from the figure that the number of instances in the neutral class is significantly lower compared to both the positive and negative classes. This indicates that the dataset is imbalanced.\par

\begin{table}[h!]
\centering
    \captionsetup{justification=centering}
    \caption{Bengali and English Caption Count}\label{capt}
    \begin{tabular}{|l|l|lll}
    \cline{1-2}
    Language & Caption Count &  &  &  \\ \cline{1-2}
    Bengali  & 2511           &  &  &  \\ \cline{1-2}
    English  & 451            &  &  &  \\ \cline{1-2}
    Bengali + English    & 1389           &  &  &  \\ \cline{1-2}
    No Caption    & 17           &  &  &  \\ \cline{1-2}
    \end{tabular}
\end{table}

The MemoSEN dataset contains captions of multiple language, including Bengali, English, and a combination of both languages. Statistics of caption language is shown in Table \ref{capt}.

\begin{figure}[htbp]
    \centerline{\includegraphics[width=\columnwidth]{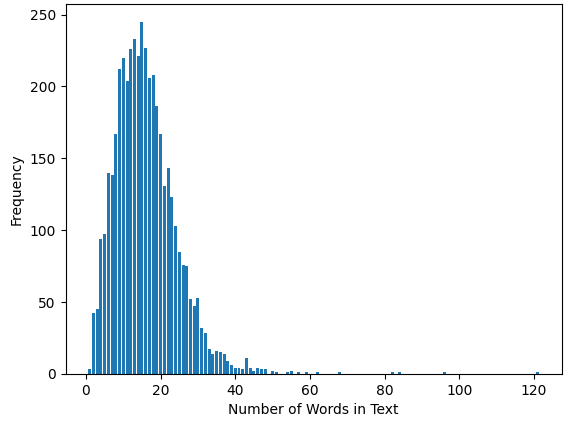}}
    \caption{Length-Frequency distribution of Captions}
    \label{fig:LenFreq}
\end{figure}

\begin{figure}[htbp]
    \centerline{\includegraphics[width=\columnwidth, height=5cm]{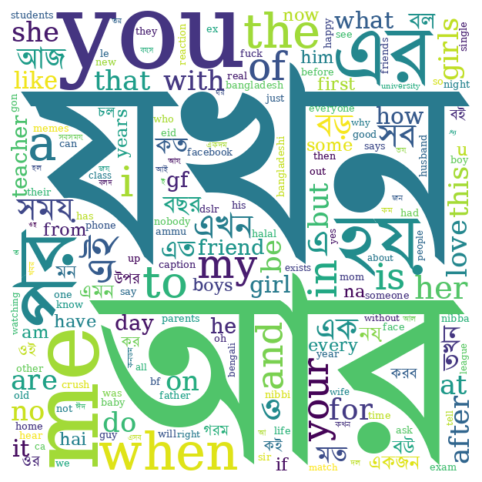}}
    \caption{Word Cloud}\label{ntype}
    \label{fig:WordCloud}
\end{figure}

Captions in this dataset vary in length. The longest caption comprises 121 words, whereas the shortest contains just 1 word. On average, captions consist of approximately 19 words. The
length-frequency distribution over the dataset is visualized in Figure \ref{fig:LenFreq}. Also, a word cloud is shown in Figure \ref{fig:WordCloud} as an addition to data analysis to visually represent the top 200 frequently used words taken from the captions.

\section{Methodology}

\begin{figure}[h!]
    \centerline{\includegraphics[width=\columnwidth, height=9.2cm]{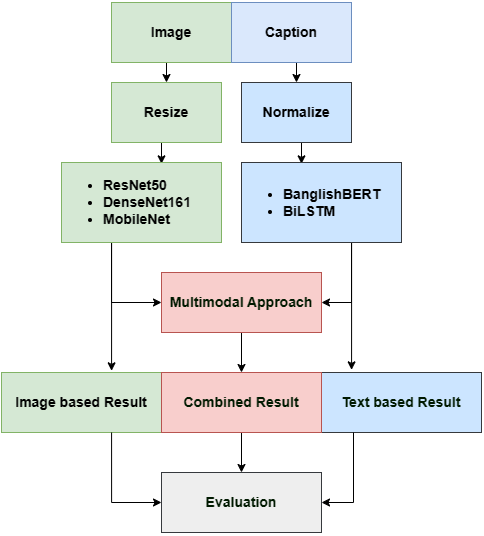}}
    \caption{Methodology}\label{ntype}
    \label{fig:meth}
\end{figure}

The methodology followed in the study is depicted in Figure \ref{fig:meth}. An in-depth description of the steps is provided in this section.

\subsection{Dataset Preparation} In the dataset, we encountered instances where some of the image path entries did not correspond to valid image files. As a result, we decided to remove those entries from the dataset.

\subsection{Data Preprocessing} The dataset consists of multimodal data, including both images and texts. To ensure clarity and meaningful information, preprocessing steps were applied to both modalities. For the preprocessing, the following steps were performed: 
\begin{itemize} 
\item \textbf{Normalizing:} Punctuations and characters in the captions that
have multiple unicode representations were normalized using a normalizer module \cite{hasan-etal-2020-low} to reduce data sparsity. 
\item \textbf{Resizing:} Images were resized to a resolution of 224x224 pixels to ensure consistency in image shape. 
\end{itemize}

\subsection{Classification} For classification, we employed visual, textual, and multimodal approaches individually. The models that were used are listed as follows:
\begin{itemize} 
\item \textbf{Visual/Image-based Approach:} We used ResNet50 \cite{he2016deep}, MobileNet v3 \cite{howard2019searching}, and DenseNet161 \cite{huang2017densely} to classify the memes with only the images as input. All of these models were initilized with ImageNet1k \cite{imagenet15russakovsky} weights and then trained on MemoSen dataset.
\item \textbf{Textual/Text-based Approach:} As the captions include both Bengali and English words, this presents a challenge where state-of-the-art BERT models trained on a single language will perform poorly for downstream tasks. For proper classification based on textual data, a multilingual BERT was necessary. For this reason, we employed BanglishBERT \cite{bhattacharjee-etal-2022-banglabert}, an ELECTRA discriminator model pre-trained with the Replaced Token Detection (RTD) objective on large amounts of Bengali and English corpora. Additionally, we employed BiLSTM for a performance comparison with BanglishBERT.
\item \textbf{Multimodal Approach:} For the multimodal approach, we tried combinations of BanglishBERT + ResNet50 and BanglishBert + DenseNet161. First, each models were used to extract features. The features were then passed through two different linear layers each containing 20 neurons. The output of the linear layers was concatenated and passed through another linear layer with 3 neurons.
\end{itemize}

\subsection{Evaluation:} We evaluated all of our approaches with accuracy, precision, recall, and f1-score. As the dataset is imbalanced, weighted scores were used.

\section{Experimental Result}
% \usepackage{multirow}
\begin{table}[H]
\captionsetup{justification=centering}
\caption{Performance Comparison of Visual, Textual and Multimodal Approaches}\label{41}
\begin{tabular}{p{0.05\linewidth}p{0.18\linewidth}|l|l|l|p{0.1\linewidth}|}
\cline{3-6}
                                                                                                &                                                                & \textbf{Accuracy} & \textbf{Precision} & \textbf{Recall} & \textbf{F1-Score} \\ \hline
\multicolumn{1}{|l|}{\multirow{3}{*}{Visual}}                                                    & ResNet50                                                      & 0.72              & 0.67               & 0.72            & 0.69              \\ \cline{2-6} 
\multicolumn{1}{|l|}{}                                                                          & \begin{tabular}[c]{@{}l@{}}MobileNet v3\\ (Large)\end{tabular} & 0.71              & 0.66               & 0.71            & 0.67              \\ \cline{2-6} 
\multicolumn{1}{|l|}{}                                                                          & DenseNet161                                                   & 0.73              & 0.69               & 0.73            & 0.70              \\ \hline
\multicolumn{1}{|l|}{\multirow{2}{*}{Textual}}                                                     & BiLSTM                                                         & 0.62              & 0.39               & 0.62            & 0.48              \\ \cline{2-6} 
\multicolumn{1}{|l|}{}                                                                          & BanglishBERT                                                   & 0.66              & 0.66               & 0.66            & 0.66              \\ \hline
\multicolumn{1}{|l|}{\multirow{2}{*}{\begin{tabular}[c]{@{}l@{}}Visual\\ +\\ Textual\end{tabular}}} & \begin{tabular}[c]{@{}l@{}}BanglishBERT\\+ DenseNet161\end{tabular}   & 0.73              & 0.68               & 0.73            & 0.70              \\ \cline{2-6} 
\multicolumn{1}{|l|}{}                                                                          & \begin{tabular}[c]{@{}l@{}}Banglish BERT\\+ ResNet50\end{tabular}      & \textbf{0.74}     & \textbf{0.69}      & \textbf{0.74}   & \textbf{0.71}     \\ \hline
\end{tabular}
\end{table}

Table \ref{41} shows the weighted scores of all the models we have tested. As we can see, the visual approaches outperform the textual approaches in every case, since the lowest F1-score achieved by MobileNet v3 is 0.67, whereas the highest F1-score in the textual approaches was achieved by BanglishBERT with only 0.66. The best result overall is obtained by combining the visual and textual approaches of BanglishBERT and ResNet50, which achieved a 0.71 weighted F1-score. This implementation outperforms the previously attained highest F1-Score of 0.635 with ResNet50+BiLSTM, as reported
by Hossain et al. \cite{hossain2022memosen}.

\begin{figure}[h]
    \centerline{\includegraphics[width=\columnwidth]{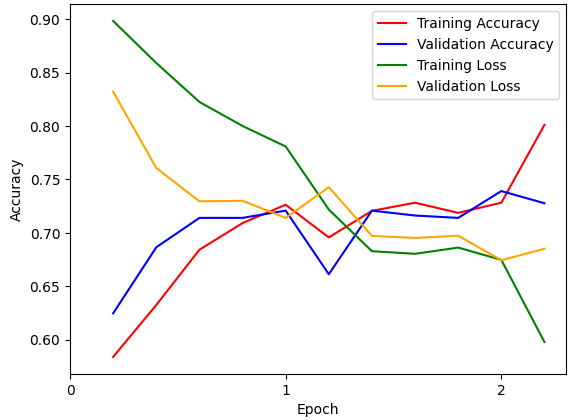}}
    \caption{Train and Validation Loss vs Accuracy Curve of Multimodal BanglishBERT + ResNet50}
    \label{fig:mtlaccloss}
\end{figure}

For a better understanding of the training process, the train and validation loss vs. accuracy graph of our multimodal approach is visualized in Figure \ref{fig:mtlaccloss}. To minimize bias and variance of the model, different hyperparameters were tested. The best result was achieved using a batch size of 32 and the AdamW optimizer with a learning rate of 0.00001, betas of (0.9, 0.9999), an epsilon value of 1e-9, and a weight decay of 0.08.

\begin{figure*}[t]
    \centering
    \begin{subfigure}[b]{0.65\columnwidth}
    \centering
        \includegraphics[width=\columnwidth]{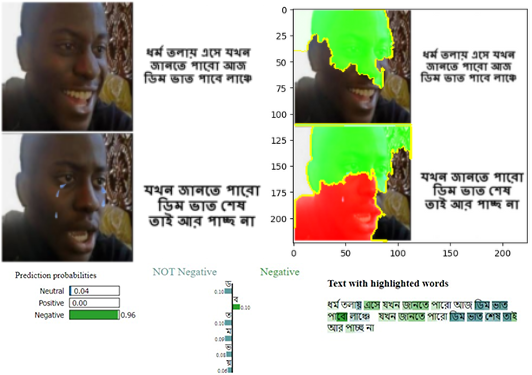}\label{nn1}
        \subcaption[]{True Label: Neutral, Textual: Negative,\\Visual: Negative, Multimodal: Negative\\}
    \end{subfigure}
    \begin{subfigure}[b]{0.65\columnwidth}
        \centering
        \includegraphics[width=\columnwidth]{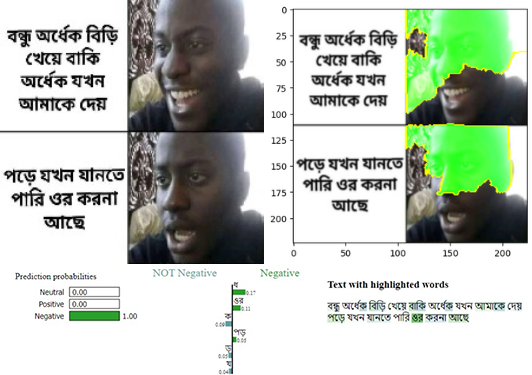}\label{nn2}
        \subcaption[]{True Label: Negative, Textual: Negative,\\Visual: Negative, Multimodal: Negative\\}
    \end{subfigure}
    \centering
    \begin{subfigure}[b]{0.65\columnwidth}
    \centering
        \includegraphics[width=\columnwidth]{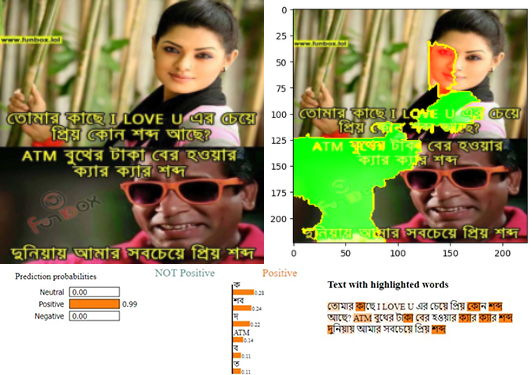}
        \subcaption[]{True Label: Positive, Textual: Positive,\\Visual: Positive, Multimodal: Positive\\}
    \end{subfigure}
    \hfill
    \begin{subfigure}[b]{0.65\columnwidth}
        \centering
        \includegraphics[width=\columnwidth]{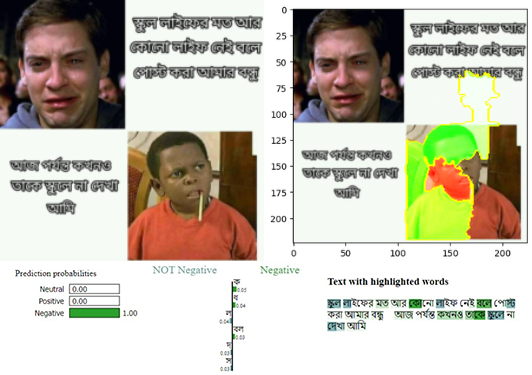}
        \subcaption[]{True Label: Positive, Textual: Negative,\\Visual: Negative, Multimodal: Negative\\}
    \end{subfigure}
    \centering
    \begin{subfigure}[b]{0.65\columnwidth}
    \centering
        \includegraphics[width=\columnwidth]{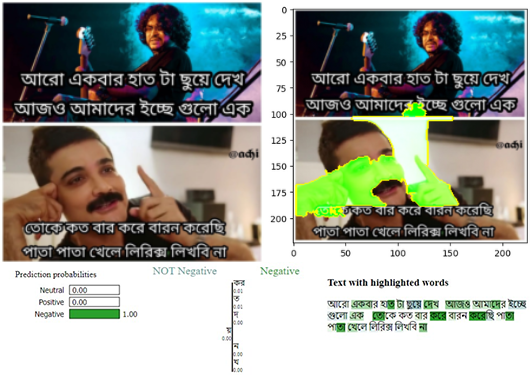}
        \subcaption[]{True Label: Negative, Textual: Negative,\\Visual: Negative, Multimodal: Negative\\}
    \end{subfigure}
    \begin{subfigure}[b]{0.65\columnwidth}
        \centering
        \includegraphics[width=\columnwidth]{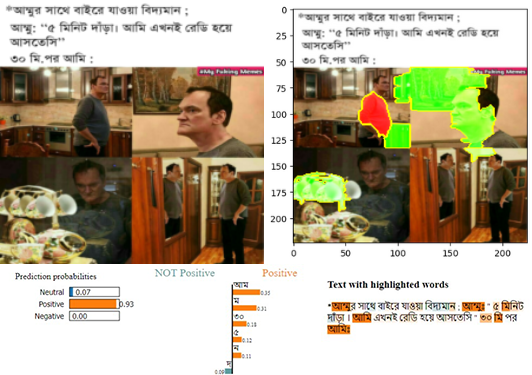}
        \subcaption[]{True Label: Negative, Textual: Positive,\\Visual: Negative, Multimodal: Positive\\}
    \end{subfigure}
    \caption{XAI implemented example of memes}
    \label{fig:xai}
\end{figure*}

\begin{figure}[H]
    \centering
    \begin{subfigure}[b]{0.67\columnwidth}
        \includegraphics[width=\columnwidth]{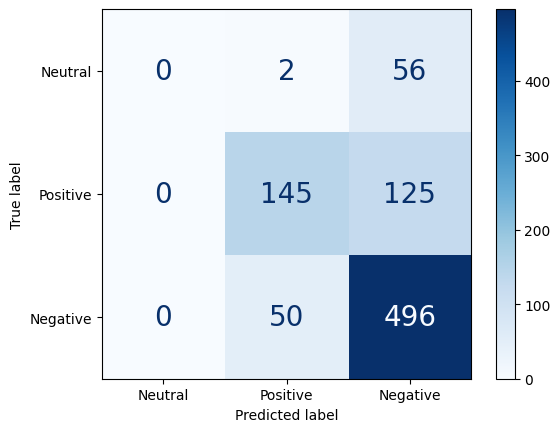}
        \subcaption[]{DenseNet161}
    \end{subfigure}
    \begin{subfigure}[b]{0.67\columnwidth}
        \includegraphics[width=\columnwidth]{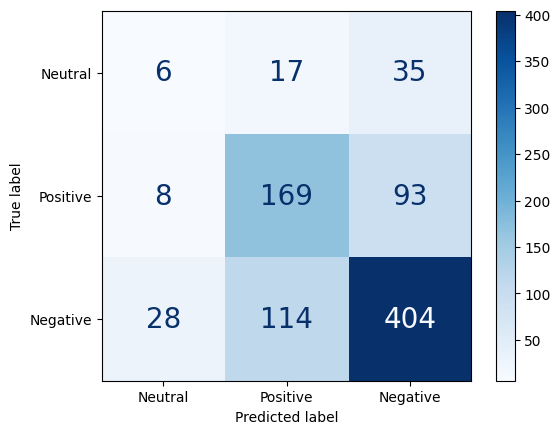}
        \subcaption[]{BanglishBERT}
    \end{subfigure}
    \begin{subfigure}[b]{0.67\columnwidth}
        \includegraphics[width=\columnwidth]{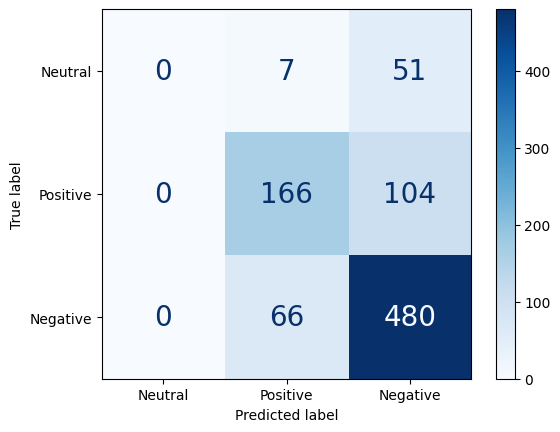}
        \subcaption[]{BanglishBERT + ResNet50 }
    \end{subfigure}
    \caption{Confusion Matrix of highest performing models}
    \label{fig:CM}
\end{figure}

Figure \ref{fig:CM} illustrates the confusion matrix of the best-performing models of Visual, Textual, and Multimodal approaches. A striking observation is that none of the models could accurately classify neutral memes. Both the Visual and Multimodal approaches failed to correctly identify any neutral memes, while the Textual approach only managed to classify six neutral memes correctly. This suggests that neutral memes are more challenging to detect than positive or negative memes and require more sophisticated methods to capture their subtleties.

\section{Result Analysis}

To understand how these models make their decisions, we applied the LIME algorithm to both DenseNet161 and BanglishBERT. Figure \ref{fig:xai} visualizes six memes on which LIME was applied. The examples are organized in the following order: the original meme on the left, the result of LIME on image input on the right, and the result of LIME on captions below them. For image data, green-marked areas support the decision of the model, and red-marked areas contribute against that decision. Similarly, for textual data, orange tokens contribute to positive, green tokens contribute to negative, and blue tokens contribute to neutral sentiment. The intensity of color represents the importance of the marked token.\par
From the examples of Figure \ref{fig:xai}, it can be seen that DenseNet161 focuses mostly on people’s faces, which is not enough to understand the underlying sentiment of the memes. For proper sentiment prediction, captions are also necessary. For this reason, multimodal achieves the highest score as it uses both caption and image as its input. However, DenseNet161 was able to reach a very close score (0.70 F1-Score) compared to BanglishBERT (0.66 F1-Score) which implies that memes on the dataset rely more on images for sentimental context.

Our results reveal a common trend among all the models: none of them were able to accurately classify neutral memes. Possible reasons for this anomaly are the following:
\begin{enumerate}
    \item The number of neutral memes in the dataset is relatively low, with only 291 instances.
    \item Visually similar meme templates are used across all classes of memes, and due to the imbalanced nature of the dataset, the model may shift its focus toward other classes.
    \item Textual tokens used in neutral memes are also common in other classes.
\end{enumerate}

As evidence to support our second hypothesis, the LIME algorithm was implemented on neutral (Figure 7a) and negative (Figure 7b) memes that share the same visual template. It can be observed that, in both cases, the model focuses on the upper portion of the person’s head, which is expected since the images are identical. However, the nature of this meme is defined by its caption, in which the visual model falls short.\par

Moreover, in the text data, as illustrated in Figure 7a, the majority of the highlighted tokens are associated with the negative class, supporting our third hypothesis.

\section{Conclusion}
This study presents a comprehensive analysis of the performance of various models, using both text and image-based inputs, for Bengali meme classification. However, none of the models were able to accurately classify neutral memes, resulting in poor performance scores. To further advance this study, more data could be added to balance the dataset, and state-of-the-art VisualBERT-based models could be implemented. Our contribution to this study includes a thorough analysis of meme classification using unimodal (Image, Text) and multimodal (Image + Text) approaches, reaching a satisfactory performance with the multimodal approach, as well as an in-depth explanation of anomalies through the implementation of LIME.

\bibliographystyle{IEEEtran}
\bibliography{bibtex} 

\end{document}